\pdfoutput=1

\documentclass[11pt]{article}

\usepackage{acl}

\usepackage{times}
\usepackage{latexsym}

\usepackage[T1]{fontenc}

\usepackage[utf8]{inputenc}

\usepackage{microtype}

\usepackage{inconsolata}
\usepackage{graphicx} 
\usepackage{algorithm}
\usepackage{algorithmic}
\usepackage{amsmath} 
\usepackage{amsfonts}
\usepackage{newfloat}
\usepackage{listings}
\usepackage{multirow}
\usepackage{makecell}
\usepackage{subfig}
\usepackage{dsfont} 
\usepackage[most]{tcolorbox}
\usepackage{hyperref}
\usepackage{xcolor}
\title{MSI-Agent: Incorporating Multi-Scale Insight into Embodied Agents for Superior Planning and Decision-Making}

\author{Dayuan Fu$^{1,2}$, Biqing Qi$^{1,3}$\thanks{Corresponding authors.}, Yihuai Gao$^{4}$\thanks{Yihuai Gao participated in this project when he was an undergrad in Tsinghua University}, \bf Che Jiang$^{1}$,  Guanting Dong$^{2}$, Bowen Zhou$^{1,3}$\footnotemark[1] \\ 
  $^1$Department of Electronic Engineering, Tsinghua University\\
$^{2}$ Beijing University of Posts and Telecommunications, Beijing, China \\
$^{3}$Shanghai AI Laboratory \\
$^{4}$Stanford University \\
  \texttt{fdy@bupt.edu.cn}\\
  \texttt{zhoubowen@tsinghua.edu.cn}
  }

\begin{document}
\maketitle
\begin{abstract}

Long-term memory is significant for agents, in which insights play a crucial role. However, the emergence of irrelevant insight and the lack of general insight can greatly undermine the effectiveness of insight.
To solve this problem, in this paper, we introduce \textbf{M}ulti-\textbf{S}cale \textbf{I}nsight Agent (MSI-Agent), an embodied agent designed to improve LLMs' planning and decision-making ability by summarizing and utilizing insight effectively across different scales. MSI achieves this through the experience selector, insight generator, and insight selector. Leveraging a three-part pipeline, MSI can generate task-specific and high-level insight, store it in a database, and then use relevant insight from it to aid in decision-making.
Our experiments show that MSI outperforms another insight strategy when planning by GPT3.5. 
Moreover, We delve into the strategies for selecting seed experience and insight, aiming to provide LLM with more useful and relevant insight for better decision-making.
Our observations also indicate that MSI exhibits better robustness when facing domain-shifting scenarios. 

\end{abstract}

\section{Introduction}
%

\begin{figure}[t]
  \centering
  \includegraphics[width=0.47 \textwidth]{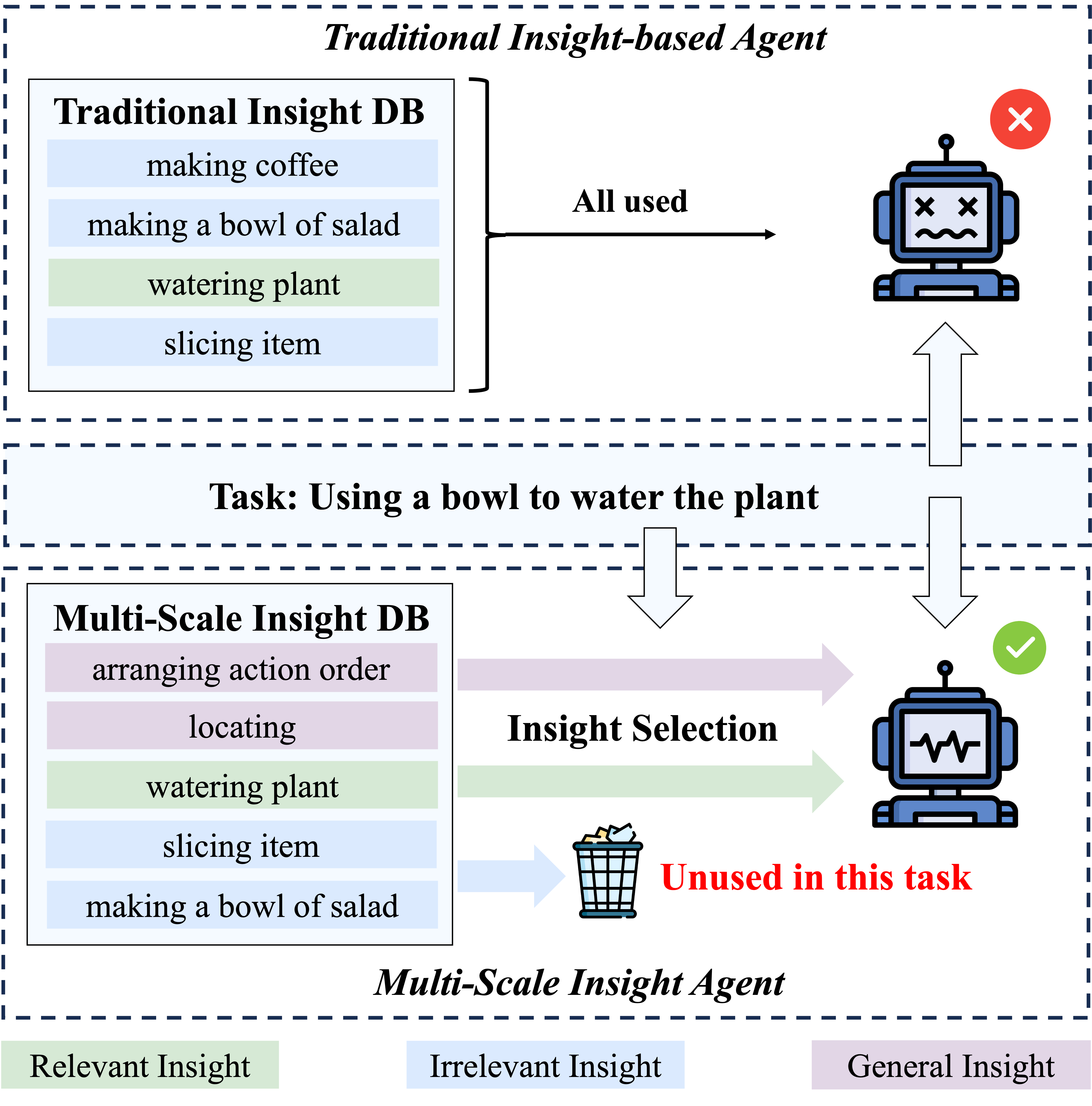}
  \caption{Example of insight summarizing and utilizing. MSI will summarize the insights in multi-scale and utilize insights by selecting based on the task. DB=Database.}
  \label{fig:example1}
\end{figure}

Creating agents that can make autonomous decisions in the environment has always been a promising and interesting research direction. \cite{AutoGPT,sun2023adaplanner} 
With the emergence of ChatGPT and GPT-4 \cite{achiam2023gpt}, large language models (LLMs) 
have transformed from specialized models to a general model that can complete multiple types of tasks, hence it can make decisions for agents. \cite{xi2023rise,yang2024qwen2,wang2023survey}.
This type of agent will transform multi-modal information into natural language as short-term memory. It then prompts large language models with short-term memory and long-term memory to plan and make decisions. With these capabilities, the agent can generate a series of actions that are executable within a given environment. \cite{yao2023tree,park2023generative,gao2023strategyllm,zheng2023synapse}

Insight\footnote{In this paper, "insight" refers to "the knowledge acquired through multiple observations of facts or events"}, as a form of long-term memory, has gradually become a crucial part of guiding LLM planning and decision-making. \cite{shinn2023reflexion,zhao2023expel,DBLP:journals/corr/abs-2402-11534,wang2023voyager,xi2023rise,zeng2024automatic}.
Relative to other long-term memory such as examples, insight is more concise and higher-level.
Although previous work has proposed a method of using LLM to summarize and utilize insights \cite{zhao2023expel}, it either provides LLM with too many irrelevant insights or can not summarize the high-level insights, as shown in Figure \ref{fig:example1}. The former can \textbf{interfere with decision-making} \cite{liu2023lost,chen2023benchmarking,ren2023investigating,10.1145/3583780.3615150}, while the latter may result in \textbf{a lack of high-level prior information to assist in decision-making}. \cite{wen2023dilu,majumder2023clin,wang2023learning}. Therefore, providing models with comprehensive and related insights to the current task has become important.

\begin{figure}[t]
  \centering
  \includegraphics[width=0.47 \textwidth]{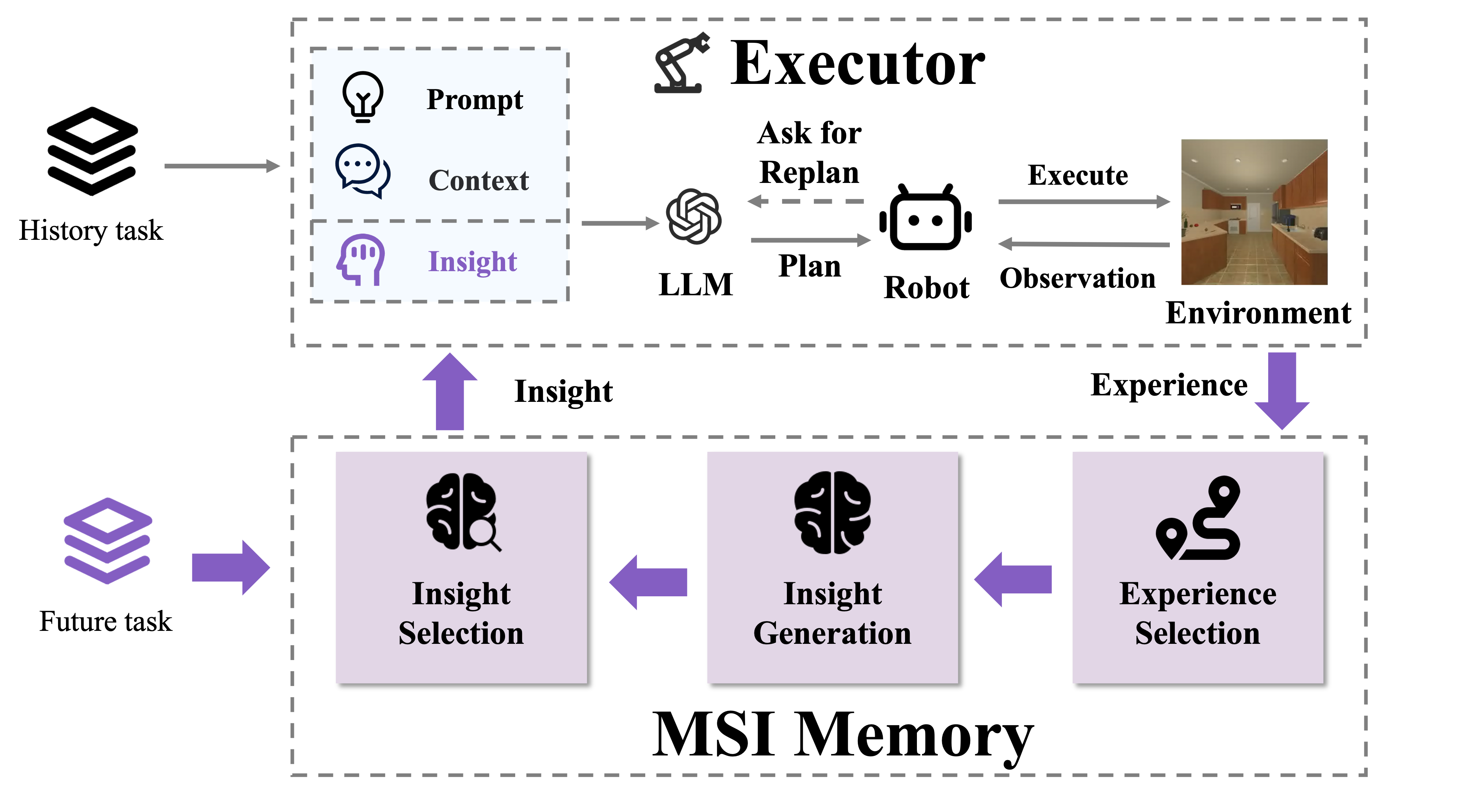}
  \caption{The overall pipeline for the MSI-agent to complete a task. MSI Memory refers to the part that deals with insight. In MSI Memory, Experience Selection and Insight Generation will summarize historical experience into insights, while Insight Selection will select insights to assist the executor in completing future tasks.}
  \label{fig:pipeline}
\end{figure}

\begin{figure*}[t]
  \centering
  \includegraphics[width=0.97 \textwidth]{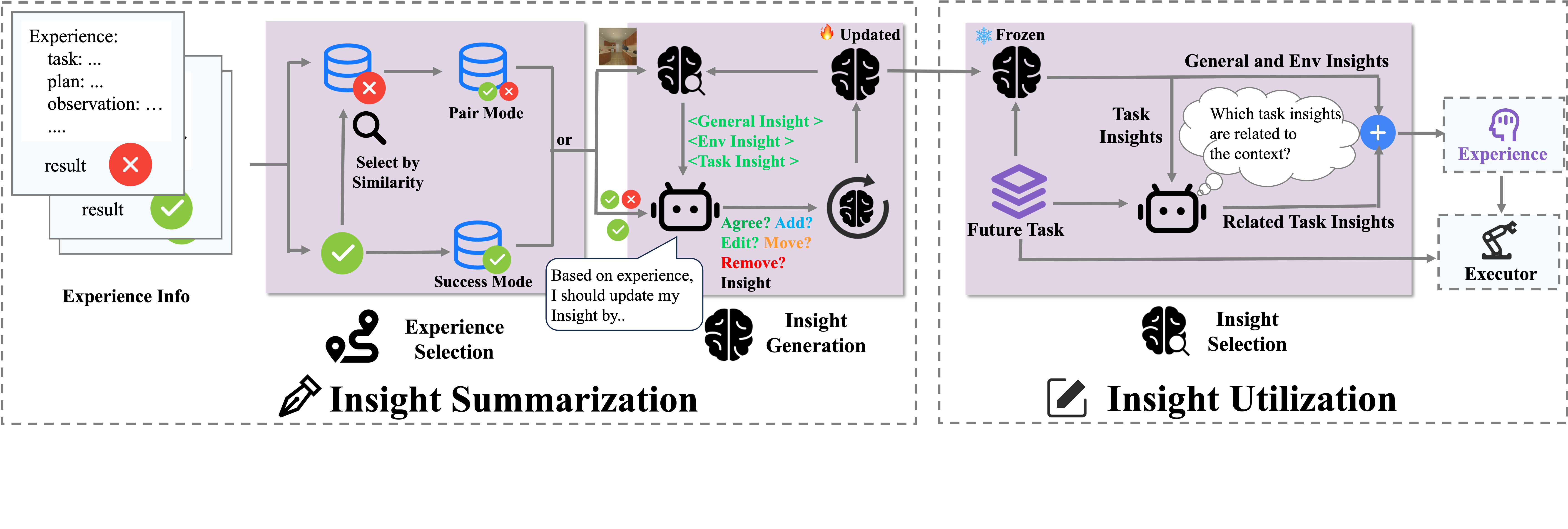}
  \caption{Pipeline of MSI Memory. The Insight Summarization part will summarize the historical task experience, while the Insight Utilization part will select relative insights to help the agent decide on future work. In the Insight Generation part, we will continuously update the insight database based on the training task experience (pair). We will freeze the database after updating insight with all training tasks. It should be noted that only some task generates environment insights (aligning with \S \ref{defination}). Env=environment}
  \label{fig:model_pipeline}
\end{figure*}

To address these challenges, we proposed \textbf{M}ulti-\textbf{S}cale \textbf{I}nsight Agent (\textbf{MSI-Agent}), an embodied agent designed to summarize and utilize insights effectively. Inspired by Expel \cite{zhao2023expel}, 
MSI collects the task background, user queries, agent's plans, environmental feedback, and execution results as \textbf{"experience"} from a series of training tasks.
 These experiences are then organized into the successful experience set or success-failure experience pairs set via an experience selector. Subsequently, an insight generator summarizes multi-scale insights based on the organized experience(s). Through this method, \textbf{both high-level and fine-grained insight can be generated}.


During task execution, the insight will pass an insight selector \textbf{to filter out the irrelevant insight} and the remaining insight prompts the executor to formulate plans and execute tasks within a given environment. The overall pipeline for the MSI agent to complete a task is illustrated in Figure \ref{fig:pipeline}, while the architecture of the insight part in MSI is detailed in Figure \ref{fig:model_pipeline}.

This solution effectively mitigates the issues highlighted earlier. By allowing classifying and selecting insights, MSI ensures that the LLM is not overwhelmed with irrelevant insights. Simultaneously, the multi-scale insights generation provides a nuanced understanding at various levels, addressing the challenge of high-level insights summarization. As a result, MSI stands as a robust solution, offering contextual and comprehensive insights tailored to enhance decision-making capabilities.

In summary, our contributions are as follows:

(1) We proposed MSI, an embodied agent that can create and utilize multiple scales of insights, greatly improving the alignment between insights and tasks.

(2) We designed 3 useful modules among experience selection, multi-scale insight generation, and task-related insight selection, shielding the noise caused by irrelevant insights.

(3) We got the SOTA results in the TEACh TfD benchmark with GPT3.5 and beat another insight mechanism in the Alfworld. What's more, our experiment comprehensively investigates the selection strategies of seed experiences and insights under various approaches and has proven that the MSI can enhance the robustness of insight utilization facing domain shifting.

\section{Related Work}
\subsection{Embodied AI}

 Embodied AI focuses on leveraging multi-model information for decision and execution of actions. Diverging from traditional reinforcement learning approaches \cite{schulman2017proximal}, current research endeavors employ language models as decision-makers for action decisions. Specifically, the model transforms information from non-natural language modalities into natural language through a modality transformer \cite{inoue2022prompter,sarch2023open}, using natural language information as input to guide the Large Language Model in decision-making \cite{song2023llm,singh2023progprompt,singh2022ask4help,suglia2021embodied,DBLP:journals/corr/abs-2402-11534}. Some methods involve fine-tuning the language model to map language inputs to action sequences at different hierarchical levels \cite{zhang2022danli,zheng2022jarvis,koshti2023interaction}, while others prompt a frozen LLM to predict action plans, relying on the instruction-following and context-learning properties of the LLM to simulate new tasks during testing \cite{wu2023tidybot,sarch2023open,song2023llm,singh2023progprompt,singh2022ask4help,DBLP:journals/corr/abs-2406-13542}. By relying on action(s) generated by the model, the robot can accomplish the designated tasks in the environment.
\subsection{LLM Long-term Memory} 
When making decisions, humans often recall past cases to assist in decision-making. Due to the limited input length, the LLM Agent cannot receive infinite historical experiences. Therefore, efficiently utilizing existing success/failure experiences becomes crucial. The LLM Long-term Memory is designed to address this challenging issue \cite{zhao2023expel,wen2023dilu,majumder2023clin,qian2024investigateconsolidateexploit}.
Currently, the LLM Agent Memory operates in two modes: example memory and insight memory. Example memory involves manually crafting experience examples that were successful in tasks. During usage, similar examples are retrieved based on the current task, using methods such as vectors or BM25, to prompt the large language model \cite{wang2023voyager,wen2023dilu,DBLP:journals/corr/abs-2406-18676,song2023llm,zhong2023memorybank}.
Insight memory, on the other hand, summarizes success/failure experiences into insights through the LLM. When new tasks occur, the insights are directly input as a part of the prompt into the LLM for helping planning and decision-making. \cite{majumder2023clin,zhao2023expel}.

\section{Method}
Figures \ref{fig:pipeline} and \ref{fig:model_pipeline} illustrate our approach. Initially, utilizing historical task data (train set), we employ the task execution module to collect a sufficient number of experiences. (\S \ref{task execution}) These experiences are then subjected to the experience selector, which identifies experiences/experience pairs suitable for generating insights. (\S \ref{Scheme Generation and Selection}) Subsequently, the multi-scale insights will be generated and stored in the insight database. (\S \ref{Multi-Scale Experience Generation})  When a new task arises, we retrieve relevant sights from the database based on predefined rules. (\S \ref{Multi-Scale Experience Selection}) These insights, along with task background, and user queries, are provided to the task execution module to facilitate execution. We refer to the process from experience collection to insight generation as insight summarization, and the subsequent insight selection and task execution as insight utilization.

\subsection{Experience Generation} 
\label{task execution}

As shown in Figure \ref{fig:pipeline}, we regard training data as history tasks. For each history task,
 the executor leverages LLM to generate a plan based on task background and user queries. Subsequently, the robot employs first-order logic to decompose the plan into atomic actions (e.g., moving forward, picking up objects) and execute them in an environment. In some tasks or cases, the executor may replan based on the environment feedback. Upon completion, task background, user queries, agent's plans, environmental feedback, and execution results are stored as experiences for summarization. Detailed information can be found in Appendix \ref{executor}.

\subsection{Experience Selection} 
\label{Scheme Generation and Selection}
The selection of experiences is crucial in summarizing insights, as it determines the quality of insights the model consolidates. As shown in Figure \ref{fig:model_pipeline}, our Experience Selection employs two modes:  

\textbf{Success mode:} We select experiences with successful execution results as the success mode experiences. 

\textbf{Pair mode:} 
For each successful experience $s_{s}$, we identify a corresponding experience $s_{f}$ from the unsuccessful experience database $S_{f}$ by:
\begin{equation}
    s_{f} = \operatorname*{argmax}_{s \in S_{f}} \frac{emb(s) \cdot emb(s_{s})}{\sqrt{||emb(s)||_2  ||emb(s_{s})||_2} }
\end{equation}

Where $emb$ is the embedding of the experience's user query and the $(s_{s},s_{f})$ is the final experience pair in the pair mode.


These two types of selected experience (pair) collections are subsequently preserved and utilized as seed experience for insight summarization.

\subsection{Multi-Scale Insight Generation} 
\label{Multi-Scale Experience Generation}


\textbf{Multi-Scale Insight} 
\label{defination}
We categorize the insights into several scales.
For all tasks, we will generate general scale and subtask scale insights. If the task provides a specific environment category (for example, kitchen), we will also generate environment scale insights.
General insight refers to the knowledge required for all tasks, which should be high-level. Environment insight pertains to the knowledge needed in a specific environment, and subtask insight involves the understanding of executing particular subtasks. The overall pipeline can be seen in Figure \ref{fig:model_pipeline}'s Insight Generation module.

\textbf{Insight Generation}  
We initialize the insight database to be empty. Whenever a seed experience merges, we select all insights in the order of general, subtask.\footnote{\label{3}If there is a specific environment category in the task, we will select environment and subtask insight that is consistent with the experience's environment category, and the order is general, environment, and subtask} as a pool of candidate experience for updating. 

Subsequently, we prompt the LLM with templates containing the candidate insight, all experience information, and descriptions of 5 atomic actions: adding, removing, editing, agreeing on an insight, and moving an insight between scales, requesting the LLM to update the insight database through these atomic actions \cite{zhao2023expel}. For subtask insight, we also require the LLM to additionally generate a subtask name corresponding to the insights. \footnote{The prompt of Insight Generation can be seen in Appendix \ref{generator}}

After the LLM generation is complete, we update the insight database in the order of general, environment (if have), and subtask, according to the atomic actions. 

Align with Expel, we also employ a scoring mechanism in insight generation. Specifically, each insight receives an initial score of 2 when an "add" or "move" action is executed, the score increases by 1 for an "agree" action, remains unchanged for an "edit" action, and decreases by 1 for a "remove" action. An insight is discarded when its score reaches zero.

\subsection{Multi-Scale Insight Selection} 
\label{Multi-Scale Experience Selection}





Similar to the generation process, we use general and subtask insights$^{\ref{3}}$ as candidate insights. For subtask insights, we adopt two modes for further selection:

\textbf{Hashmap indexing}: We extract all subtask names from the subtask insight database, combine them with user queries, and provide them to the LLM, requiring the LLM to return all task names related to the user query. Subsequently, we consider all insights under returned subtask names as the subtask insights for this user query. The prompt of hashmap subtask selection can be seen in Appendix \ref{hash}

\textbf{Vector indexing}: We compute the cosine similarity between all subtask insights and the user query, selecting insights with at most 2000 tokens.\footnote{Due to the excessive noise through vector indexing, we utilize this method only in ablation experiments.}

Ultimately, we provide the different scales of insights, and the user query to the task execution module to accomplish the task.
\section{Experiment}


We evaluate MSI on the 2 benchmarks\footnote{Detailed information can be seen in Appendix \ref{dataset}.}: TEACh TfD benchmark \cite{padmakumar2022teach} and AgentBench Alfworld benchmark \cite{shridhar2020alfworld,liu2023agentbench}.  Our experiments are designed to address the following research questions (RQs): \textbf{RQ1}: Does MSI outperform other insights methods? \textbf{RQ2}: What kind of seed experience selection strategy should be chosen when facing different insight generation strategies and tasks? \textbf{RQ3}: What kind of insight selection strategy should be adopted for different future tasks? \textbf{RQ4}: How does the robustness of the MSI system evolve with the domain shifts?

\subsection{Experimental Setup} 
\textbf{Evaluation metrics} For TEACh, we calculate accuracy $(ACC)$ and path length weighted $(PLW)$ metrics under two settings: Task Success Rate $(SR)$ and Goal Condition Success Rate $(GC)$. Aligned with HELPER, these four metrics are:
\begin{equation}
    SR_{ACC} = E_{x\sim p}\left(\mathds{1}(SCN_x = GCN_x)\right)
\end{equation}

\begin{equation}
    GC_{ACC} = \frac{\sum_{x\sim p}SCN_x}{\sum_{x\sim p}GCN_x} \;\;\;
\end{equation}

\begin{equation}
    SR_{PLW} = \frac{\sum_{x\sim p}\frac{\mathds{1}(SCN_x = GCN_x)*L_{ref_{x}}^2}{Max( L_{pred_{x}}, L_{ref_{x}})}}{\sum_{x\sim p}L_{ref_{x}}}
\end{equation}

\begin{equation}
    GC_{PLW} = \frac{\sum_{x\sim p}\frac{(SCN_x / GCN_x)*L_{ref_{x}}^2}{Max( L_{pred_{x}}, L_{ref_{x}})}}{\sum_{x\sim p}L_{ref_{x}}}
\end{equation}

$SCN$ and $GCN$ refer to the success condition number and goal condition number respectively, $L_{pred}$ refers to the step used to execute the task by the executor while $L_{ref}$ refers to the step used to execute the task by a human annotator, $p$ refers to the distribution of the datasets and $x$ is the sample of the distribution of the datasets.

For Alfworld, we calculate the $SR_{ACC}$ metric.


\subsection{Executor} 
 \textbf{TEACh} We use HELPER \cite{sarch2023open} as the TEACh's executor. HELPER \cite{sarch2023open} is an executor framework built on top of TEACh. As shown in Figure \ref{fig:pipeline}, it provides the task background, user query (i.e., the dialogue), and other relevant information to the LLM in a fixed format, allowing the LLM to generate a piece of code as the plan\cite{chen2021evaluating} and create a sequence of subtasks to guide the robot. Initially, the robot will walk around the environment to observe and obtain a spatial plan map that includes information about the objects it has observed, as well as its location \cite{blukis2022persistent}. At each time step, the robot receives an RGB image through its camera. It will then determine an atomic action based on the image, location, and subtask, and execute it in the simulation environment. \cite{sarch2023open,zhang2022danli} If the execution fails, the robot will call upon the VLM model \cite{li2023multimodal} to provide the most likely reason for the failure based on the image and attempt a second try or replan \cite{yao2022react,shinn2023reflexion}. In the MSI, we include the environment, dialogue, planned subtasks, actual executed subtasks, and the VLM-provided failure reasons during replanning as part of the experience. (Note that: The EXPERIENCE in the prompt refers to insight in the paper. )
 
  \textbf{Alfworld} We use AgentBench as the Alfworld's executor. AgentBench \cite{liu2023agentbench} is executor frameworks with ReAct format \cite{yao2022react}, Alfworld is one of its subtask. As shown in Figure \ref{fig:pipeline}, AgentBench provides the task background (as shown below), user query (i.e., the dialogue), and other relevant information to the LLM in a fixed format, allowing the LLM to generate a thought and an action (as the plan) in each turn. After the action's execution, the environment will give the feedback to the agent and the agent will replan another action based on feedback and new thoughts until the task is finished. In the MSI, we include the task background, user query, and all thought-action-observations in the task as the experience.
 The introduction of HELPER and AgentBench can be seen in Appendix \ref{executor}

\subsection{Hyperparameter}

Our insight generation and decision-making components are aligned with Expel. We have chosen ChatGPT (gpt-3.5-turbo-1106) as the LLM for selecting insight subtasks. GPT-4 (gpt-4-1106-preview) as the LLM for insight generation. During the experience selection phase, we use text-embedding-ada-002 to establish a vector library for failed experiences for retrieval purposes.

\textbf{TEACh} We have chosen ChatGPT (gpt-3.5-turbo-1106) as the decision-maker for planning. The settings for experience memory enhancement, PreCheck, Correction, and locator are all aligned with HELPER. Due to the time limitation and budget, we do not use GPT4 as the decision-maker for planning.

\textbf{Alfworld} We have chosen ChatGPT (gpt-3.5-turbo-1106) and GPT-4 (gpt-4-1106-preview) as the decision-maker for planning. The examples are all aligned with AgentBench. 

\begin{table*}[htbp]
\centering
    \resizebox{0.80\textwidth}{!}{
    \begin{tabular}{lcccc}
    \Xhline{1.5pt}
\multirow{1}[4]{*}{Model} & \multicolumn{2}{c}{Seen (IND)} & \multicolumn{2}{c}{Unseen (OOD)}\\
\cline{2-5}\multicolumn{1}{c}{} & SR    & GC    & SR    & GC \\
\Xhline{1.5pt}
\multicolumn{1}{l}{\textbf{\textit{Fine-Tune Based Model}}} & \multicolumn{1}{r}{} & \multicolumn{1}{r}{} & \multicolumn{1}{r}{} & \multicolumn{1}{r}{} \\
E.T.$^*$  & 0.48 (0.12) & 0.35 (0.59) & 1.02 (0.17) & 1.42 (4.82)\\
JARVIS$^*$  & 1.80 (0.30) & 3.10 (1.60) & 1.70 (0.20) & 5.40 (4.50) \\
FILM$^*$ & 2.9 (1.0) & 6.1 (2.5) & 5.5 (2.6) & 5.8 (11.6) \\
DANLI$^*$ & 7.98 (3.20) & 6.79 (6.57) & 4.97 (1.86) & \textbf{10.50} (10.27) \\
\hline
\multicolumn{1}{l}{\textbf{\textit{LLM Agent-Based Model}}} & \multicolumn{1}{r}{} & \multicolumn{1}{r}{} & \multicolumn{1}{r}{} & \multicolumn{1}{r}{} \\
HELPER$^*$ & -     & -     & 9.48 (1.21) & 10.05 (3.68) \\
\textbf{HELPER} & 8.84 (1.76) & \textbf{13.94} (7.65) & 10.62 (1.41) & 9.29 (3.95) \\
\textbf{Expel} & 8.28 (1.86) & 11.55 (7.83) & 8.99 (2.66) & 8.49 (6.02) \\
\hline
\textbf{MSI}  & \textbf{12.70} (2.60) & 13.66 (8.72) & \textbf{14.54 }(3.70) & 10.08(6.35) \\
\Xhline{1.5pt}
    \end{tabular}%
    }
    \caption{Trajectory from Dialogue (TfD) evaluation on the TEACh validation set. Trajectory length weighted metrics are included in ( parentheses ). SR = success rate. GC = goal condition success rate. The results with $^*$ come from \cite{sarch2023open}. We use ChatGPT as the LLM in LLM Agent-Based Model. We reproduce the HELPER in \textbf{HELPER} line and apply Expel in TEACh. Both \textbf{Expel} and \textbf{MSI} use pair mode to generate insight.}
      \label{tab:main_result}%
\end{table*}%

\begin{table}[htbp]
\centering
    \resizebox{0.5\textwidth}{!}{
    \begin{tabular}{lcccc}
    \Xhline{1.5pt}
\multirow{1}[4]{*}{Model} & \multicolumn{2}{c}{GPT3.5} & \multicolumn{2}{c}{GPT4}\\
\cline{2-5}\multicolumn{1}{c}{} & Dev (IND)    & Test (OOD)    & Dev (IND)    & Test (OOD) \\
\Xhline{1.5pt}
Act-Only & 0 & 6 & 65 & 66 \\
ReAct & 0 & 10 & 65 & 68 \\
Expel & \textbf{5}& 14 & 75 & 70 \\ 
MSI & \textbf{5} & \textbf{16}& \textbf{85} & \textbf{72}\\
\Xhline{1.5pt}
    \end{tabular}%
    }
    \caption{AgentBench Alfworld results. We reproduce all results via AgentBench's framework. Both Expel and MSI use pair mode to generate insights.}
      \label{tab:alfworld}%
\end{table}%

\subsection{Baseline} 

For TEACh, We consider the following baselines: 

\textbf{Fine-Tune Based Model:}
\textbf{Episodic Transformer (E.T.)} \cite{padmakumar2022teach} is an end-to-end multimodal transformer that can predict the action by language inputs like dialogue and images in the environment.
\textbf{Jarvis} \cite{zheng2022jarvis} and \textbf{FILM} \cite{min2022don} use a multimodal transformer to predict subgoals and transform them into atomic actions by rules.
\textbf{DANLI} \cite{zhang2022danli} uses an LM to encode language inputs to high-level subgoals and uses a PDDL model \cite{lamanna2021online} to transform subgoals, object states, and spatial maps into an atomic action. It also has a strategy to replan atomic action when facing errors in atomic action.

\textbf{LLM Agent-Based Model:}
\textbf{HELPER} \cite{sarch2023open} uses LLM to transform all information into a code and uses a code parser to parse the code into subgoals. 
\textbf{Expel} \cite{zhao2023expel} presents a pipeline to generate schemes and experience as long-term memory. Different from the original setting in Expel, our pair mode uses success-fail pairs between different tasks instead of between reflexion \cite{shinn2023reflexion} steps.

For Alfworld, We consider the following baselines: 
\textbf{Act-only} \cite{yao2022react}, \textbf{ReAct} \cite{yao2022react} and \textbf{Expel} \cite{zhao2023expel}




\begin{figure*}[t]
  \centering
  \includegraphics[width=1.00 \textwidth]{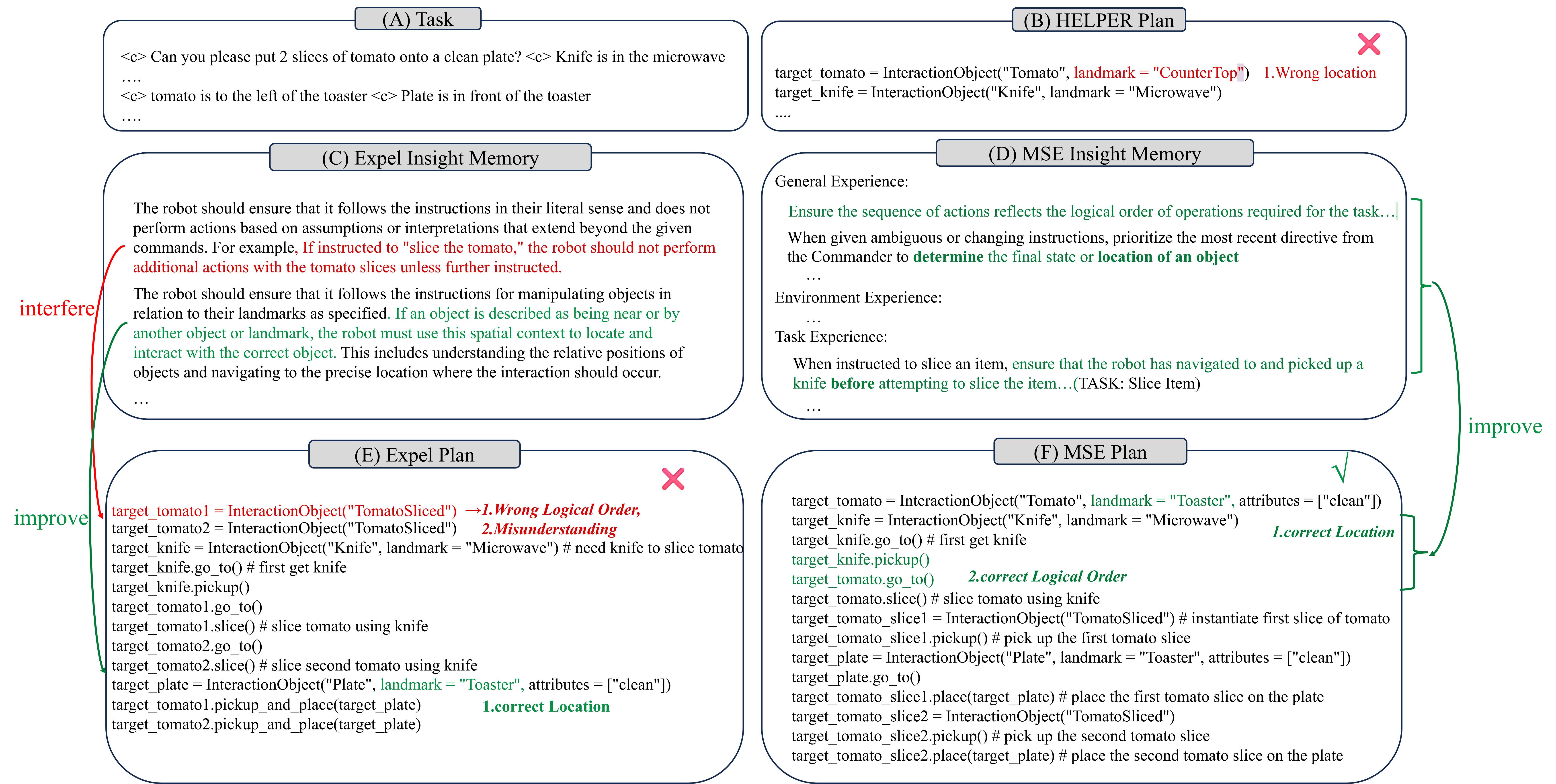}
  \caption{An example of 3 plans dealing with a specific task in TEACh. (A) The original task's user query, we omit some responses. (B) Plan to finish the task without experience. (C) Expel insights example (D) MSI insights example(E) Plan to finish the task with Expel. (F) Plan to finish the task with MSI. We omit most of the insights in Expel and MSI due to the length limitation.}
  \label{fig:example}
\end{figure*}

\subsection{Main Result (RQ1)} 

\textbf{TEACh} The performance of MSI on TEACh is displayed in Table \ref{tab:main_result}. Notably, MSI gains 12.70\% in IND data and 14.54\% in OOD data\footnote{We select only those experiences generated by GPT3.5 with  $SR_{ACC}$=1 for MSI and Expel to generate insights. Therefore, the insights should generally align with $SR_{ACC}$.}, which outperforms all results among LM and ChatGPT. In contrast, Expel performs below other LLM Agent-Based Models but above Fine-Tune Based Models. This may be because many irrelevant insights in the prompts lead to decreased performance. Despite the Expel summarizing experience based on training data, its effectiveness is inferior to that of HELPER, which uses one-shot examples directly. Conversely, MSI's success rate in both IND and OOD tasks is over 40\% higher than that of HELPER, indicating that the Multi-Scale Insight summarization and utilization method can provide task-relevant insights to assist the model in making inference decisions.

\textbf{Alfworld} The results of MSI on Alfworld are displayed in Table \ref{tab:alfworld}. Both insight mechanisms gain positive effects on ReAct-based agents. The enhancement effect on the performance through MSI insight is approximately twice that of Expel insight (20 vs 10 in GPT4-dev and 4 vs 2 in GPT4-std) which indicates the performance of MSI is meaningful over Expel.

As a result, MSI insight can improve an agent's planning and decision-making ability in both single-turn plans (TEACh) and multi-turn plans (Alfworld). This showcases its extensive versatility and potential applications across different contexts.

\textbf{Cases comparison:} Figure \ref{fig:example} illustrates the decision-making processes and insights examples used by HELPER, Expel, and MSI when completing the task of slicing tomatoes and plating them. It can be observed that HELPER incorrectly marks the landmark of Tomato as the location "CounterTop" in the one-shot example, instead of Toaster, causing a failure in finding the tomato and thus failing the task. In contrast, MSI successfully marks the landmark, even though it uses the same one-shot example where the Tomato landmark is marked as CounterTop. This is because MSI has a subtask insight that guides the model on how to ensure accurate positioning when the dialogue includes "near another object." This reflects the effectiveness of insight generation to a certain extent.

\begin{table*}[htbp]
\centering
    \resizebox{0.9\textwidth}{!}{
    \begin{tabular}{lcccccccc}
    \Xhline{1.5pt}
\multirow{2}[4]{*}{Model} & \multicolumn{4}{c}{TEACh} & \multicolumn{4}{c}{Alfworld} \\ \cline{2-9} 
& \multicolumn{2}{c}{Seen (IND)} & \multicolumn{2}{c}{Unseen (OOD)} & \multicolumn{2}{c}{Dev (IND)} & \multicolumn{2}{c}{Test (OOD)}\\
\cline{2-9}
& SR    & GC    & SR    & GC  & GPT3.5    & GPT4   & GPT3.5    & GPT4\\
\Xhline{1.5pt}
\multicolumn{1}{l}{\textbf{\textit{pair mode}}} & \multicolumn{1}{r}{} & \multicolumn{1}{r}{} & \multicolumn{1}{r}{} & \multicolumn{1}{r}{} \\
Expel & 8.28(1.86) & 11.55(7.83) & 8.99(2.66) & 8.49(6.02) & \textbf{5} & 75 & 14 &70\\
MSI   & \textbf{12.70(2.60) }& 13.66(8.72) & \textbf{14.54(3.70)} & \textbf{10.08(6.35) } & \textbf{5} &\textbf{85} & \textbf{16} & 72\\
\hline
\multicolumn{1}{l}{\textbf{\textit{success mode}}} & \multicolumn{1}{r}{} & \multicolumn{1}{r}{} & \multicolumn{1}{r}{} & \multicolumn{1}{r}{} \\
Expel & 9.94(2.25) & 11.13(7.92) & 11.60(3.04) & 9.77(6.47) & 0 & 75 & 10 & 70\\
MSI   & 10.65(1.94) & \textbf{14.15(6.69)} & 13.39(2.10) & 8,96(4.05) & 0 & 75 & 10 & \textbf{76}\\
\Xhline{1.5pt}
    \end{tabular}%
    }
    \caption{The TEACh and Alfworld result of Expel and MSI under different experience selecting strategies.}
      \label{tab:selecting}%
\end{table*}%

\begin{table*}[htbp]
\centering
    \resizebox{0.9\textwidth}{!}{
    \begin{tabular}{lcccccccc}
    \Xhline{1.5pt}
\multirow{2}[4]{*}{Model} & \multicolumn{4}{c}{TEACh} & \multicolumn{4}{c}{Alfworld} \\ \cline{2-9} 
& \multicolumn{2}{c}{Seen (IND)} & \multicolumn{2}{c}{Unseen (OOD)} & \multicolumn{2}{c}{Dev (IND)} & \multicolumn{2}{c}{Test (OOD)}\\
\cline{2-9}
& SR    & GC    & SR    & GC  & GPT3.5    & GPT4   & GPT3.5    & GPT4\\
\Xhline{1.5pt}
\multicolumn{1}{l}{\textbf{\textit{pair mode}}} & \multicolumn{1}{r}{} & \multicolumn{1}{r}{} & \multicolumn{1}{r}{} & \multicolumn{1}{r}{} \\
MSI   & \textbf{12.70(2.60)} & 13.66(8.72) & 14.54(3.70) & 10.08(6.35) & \textbf{5} &\textbf{ 85 }& 16 & 72 \\
MSI (general) & 12.15(2.36) & \textbf{13.94(8.55)} &\textbf{ 14.86(3.87)} & \textbf{11.12(7.53)} & \textbf{ 5} & 80 &\textbf{ 20} & 72\\
\hline
\multicolumn{1}{l}{\textbf{\textit{success mode}}} & \multicolumn{1}{r}{} & \multicolumn{1}{r}{} & \multicolumn{1}{r}{} & \multicolumn{1}{r}{} \\
MSI   & \textbf{10.65(1.94)} & \textbf{14.15(6.69)} & \textbf{13.39(2.10) }& 8,96(4.05) &  0 & 75 & 10 & \textbf{76} \\
MSI (general) & 10.50(2.73) & 13.66(8.87) & 12.25(3.40) & \textbf{9.81(6.17)} &  0 & 75 & 12 & \textbf{76}\\
\Xhline{1.5pt}
    \end{tabular}%
    }
    \caption{The TEACh and Alfworld result of MSI under different scale experience selecting strategies.}
      \label{tab:scale}%
\end{table*}%

Although Expel also has insight that assists the model in locating objects, and its decision-making for plate location is correct, irrelevant yet similar insight has influenced its judgment. For example, the insight marked in red in the figure may lead the LLM to mistakenly believe that it needs to generate code strictly following the dialogue sequence and that the executor needs to further slice the tomato slices. On the contrary, MSI's insight prompts the model to first determine the order of the steps, and since there are no examples in the general insight, it also reduces the LLM's susceptibility to interference from irrelevant variables.



\subsection{Experience Select Strategy (RQ2)} 

Table \ref{tab:selecting} shows the results of the two strategies under two long-term memory methods. From the perspective of the optimization goal of insights (i.e. $SR_{ACC}$), Expel performs 8.28\% and 8.99\% on HELPER IND and OOD data when using insights summarized from successful experiences alone compared to using success-failure pairs with 9.94\% and 11.60\% respectively. In contrast, MSI performs better when summarizing insights from success-failure pairs rather than just successful experiences, the former reaches 12.70\% and 14.54\% in HELPER IND and OOD data while the latter only gains 10.65\%  and 13.39\%. Alfworld's GPT3.5 version has the same trend in Table \ref{tab:selecting}. The reason for this outcome may be that Expel's method of summarizing and utilizing insights provides the LLM with many fine-grained insights that are problematic yet related to the issue or irrelevant insights(as shown in the red part of Figure \ref{fig:example}), leading to decreased accuracy. 

Conversely, when MSI summarizes the insights, it does so at multiple scales and only selects a portion for actual use by the LLM. This approach separates general insights with strong generality from fine-grained insights, ensuring that when the LLM uses insights from success-failure pairs, it can benefit from the strong generality of general insights while reducing the interference of irrelevant fine-grained insights through selective insight use. Due to this characteristic of MSI, its effectiveness in summarizing and utilizing insights from success-failure experience pairs is better than using successful experiences alone.



The above analysis indicates that the Experience Select Strategy is related to the method of generating and utilizing insights. If strong generality and specificity insights can be generated and selected, the pair mode is more helpful in enhancing the LLM's decision-making capabilities. Otherwise, the success mode should be chosen to avoid the interference of too many irrelevant insights.

\begin{table}[tbp]
\centering
    \resizebox{0.5\textwidth}{!}{
    \begin{tabular}{lcccc}
    \Xhline{1.5pt}
\multirow{1}[4]{*}{Model} & \multicolumn{2}{c}{Seen (IND)} & \multicolumn{2}{c}{Unseen (OOD)}\\
\cline{2-5}\multicolumn{1}{c}{} & SR    & GC    & SR    & GC \\
\Xhline{1.5pt}
MSI (Hashmap) & \textbf{12.70(2.60) }& \textbf{13.66(8.72) }& \textbf{14.54(3.70) }& \textbf{10.08(6.35)} \\
MSI (Vector) &  10.05(2.89) & 13.52(9.11) & 11.43(1.28) & 9.2(3.53) \\

\Xhline{1.5pt}
    \end{tabular}%
    }
    \caption{The TEACh result of MSI under different subtask insights selecting strategies.}
      \label{tab:task}%
\end{table}%

\subsection{Insights Select Strategy (RQ3)} 

Table \ref{tab:scale} shows the comparison of multi-scale insights versus only general insights used under two different Insight Select Strategies. In most cases, the use of multi-scale insights provides a stronger improvement to LLM planning than the use of general insights alone. However, when dealing with OOD problems in pair mode, the general insights gain 14.86\% in TEACh and 20\% in Alfworld, which outperforms the multi-scale insights' result of 14.54\% and 16\% respectively. This may be due to task-specific insights summarized in-domain not aligning with OOD tasks, resulting in fine-grained mismatches. Pair mode is more susceptible to fine-grained mismatches, which is why using only general insights can be more helpful to model decision-making than using multi-scale insights. Consistent with the conclusions of Section 4.4, the effectiveness of MSI when summarizing insights in pair mode is always better than in success mode.

Table \ref{tab:task} presents the impact of two different methods of refining task-specific insights on LLM decision-making in TEACh. Across both data types, results using hashmap pair retrieval are over 20\% higher on Success Rate (SR) than those using vector similarity retrieval (from 10.05\% to 12.70\% in IND and 11.43\% to 14.54\% in OOD). This is because vector similarity retrieval may introduce irrelevant insights, as shown in Figure \ref{fig:example1}. If the task is "water plants with a bowl", the top three insights retrieved by vector similarity are classified as "Water Plant", "Retrieve and Prepare" and "Prepare Beverage". The first two seem to align with the task requirements, while the third is unrelated. The "Prepare Beverage" can be retrieved because the word 'bowl' is in the task whose semantic space is associated with cooking, leading to the retrieval of irrelevant insights. This also explains why the method of vector similarity retrieval, used to retrieve schemes as examples, cannot be employed when utilizing insight.

The results from Tables \ref{tab:scale} and \ref{tab:task} collectively illustrate the strategy for selecting insight: 
 
 The agent system needs to first determine whether the current task aligns with the seed task experiences for insight generation. If there is no alignment, then only general insights in the MSI should be used to assist LLM decision-making. Conversely, if there is alignment, multi-scale insights should be used in conjunction with a key-value pair indexing strategy for selection.

\begin{figure}[t]
  \centering
  \includegraphics[width=0.5 \textwidth]{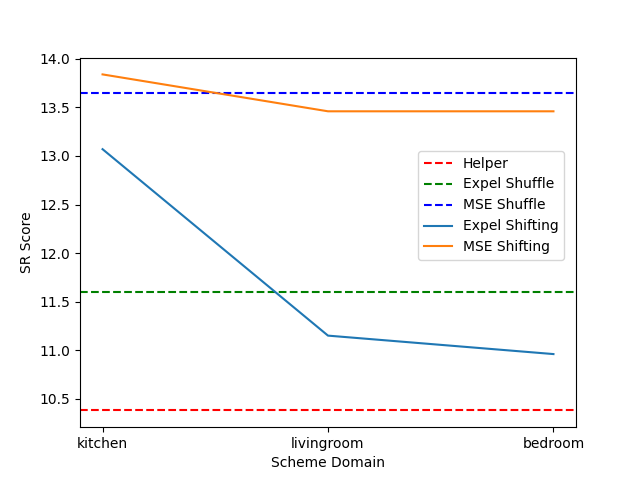}
  \caption{The robustness of agents when facing domain shifting. Dashed lines indicate baseline scores without insight or with random scheme shuffling across three domains. Solid lines show scores after sequential insight summarization: first, kitchen experiences inform insight; then living room experiences update it; finally, bedroom experiences refine it, with corresponding results displayed under each domain.}
  \label{fig:robustness}
\end{figure} 

\subsection{Robustness in Domain Adaptation (RQ4)} 
 Agents can adjust to new environments by constantly updating their insights repository. However, the distribution of new tasks may differ from that of old tasks that have already been summarized into insights, which can lead to "catastrophic forgetting" of old tasks when the insights undergo domain transfer, resulting in decreased model performance on old tasks. Therefore, it is crucial to have robust agents for Domain Adaptation.


Figure \ref{fig:robustness} illustrates the robustness of MSI and Expel under domain shifting in TEACh. We fed the training data into the insight summarizer in the order of environments: kitchen, living room, and bedroom, unlike the original MSI and Expel, which shuffle the training data before input. We selected the kitchen task in the valid unseen set as "original domain tasks" for testing. insights summarized solely on kitchen data are more beneficial in assisting the model with decision-making in the kitchen. However, as new OOD data is introduced, the model insights a degree of forgetting, leading to a decline in performance on kitchen tasks. Compared to Expel, which declines 2.11\% after summarizing the living room and bedroom scheme, MSI shows a smaller degree of performance decline and faster convergence with only a decline of about 0.38\%, proving that MSI possesses better robustness in handling domain transfer.

\subsection{Conclusion} 

In this paper, we propose MSI, which is capable of summarizing and utilizing multi-scale insights to enhance the decision-making ability of embodied agents. MSI can assist agents in making higher-quality decisions and is better equipped to handle insight distribution shifting that may occur with continuous insight updating. 

Our experiments demonstrate that for MSI, success-failure experience pairs are better seed data for insights, while the strategy for insight selection needs to be determined based on a comprehensive assessment of the future task distribution and the distribution of tasks for which insights have been summarized. 

It sets a new state-of-the-art result for the TEACh using agents based on ChatGPT as the foundation and beat another insight mechanism in the Alfworld. We believe our work contributes new insights into the summarization, storage, and utilization of long-term memory, especially insights.

\section*{Acknowledgement}
This work is supported by the National Science and Technology Major Project
(2023ZD0121403). We extend our gratitude to the anonymous reviewers for their insightful
feedback, which has greatly contributed to the improvement of this paper.

\section*{Limitations}
While MSI achieves significant improvements over existing baselines, there are still directions to explore for future work.

(1) Although the General and Subtask scale can be used in all tasks, the environment scale can only be used in some embodied scenarios. In the future, we will expand the idea of multi-scale insight by designing different scales in other tasks. 

(2) We only explore one type of long-term memory, insight. In the future, we will explore the combination of different types of long-term memory.

\bibliography{custom}

\appendix
\section{Executor}
\label{executor}
(Note that: The EXPERIENCE in the prompt refers to insight in the paper. )
\begin{tcolorbox}[
colback=white!10!white,
colframe=black!75!black,
title=HELPER executor
prompt,
breakable]
\label{response-aug prompt}
You are an adept at translating human dialogues into sequences of actions for household robots. Given a dialogue between a <Driver> and a <Commander>, you should write a Python program to be executed by a household robot that could finish all tasks in the conversation.

\{API\}

Write a script using Python and the InteractionObject class and functions defined above that could be executed by a household robot. 

Experience you have summarized in the past:
\\
\{EXPERIENCE\}
\\
\{RETRIEVED\_EXAMPLES\}

Adhere to these stringent guidelines:\\
1. Use only the classes and functions defined previously. Do not create functions that are not provided above.\\
2. Make sure that you output a consistent plan. For example, opening of the same object should not occur in successive steps.\\
3. Make sure the output is consistent with the proper affordances of objects. For example, a couch cannot be opened, so your output should never include the open() function for this object, but a fridge can be opened.\\ 
4. The input is dialogue between <Driver> and <Commander>. Interpret the dialogue into robot actions. Do not output any dialogue.\\
5. Object categories should only be chosen from the following classes: ShowerDoor, Cabinet, CounterTop, Sink, Towel, HandTowel, TowelHolder, SoapBar, ToiletPaper, ToiletPaperHanger, HandTowelHolder, SoapBottle, GarbageCan, Candle, ScrubBrush, Plunger, SinkBasin, Cloth, SprayBottle, Toilet, Faucet, ShowerHead, Box, Bed, Book, DeskLamp, BasketBall, Pen, Pillow, Pencil, CellPhone, KeyChain, Painting, CreditCard, AlarmClock, CD, Laptop, Drawer, SideTable, Chair, Blinds, Desk, Curtains, Dresser, Watch, Television, Newspaper, FloorLamp, RemoteControl, HousePlant, Statue, Ottoman, ArmChair, Sofa, DogBed, BaseballBat, TennisRacket, VacuumCleaner, Mug, ShelvingUnit, Shelf, StoveBurner, Apple, Lettuce, Bottle, Egg, Microwave, CoffeeMachine, Fork, Fridge, WineBottle, Spatula, Bread, Tomato, Pan, Cup, Pot, SaltShaker, Potato, PepperShaker, ButterKnife, StoveKnob, Toaster, DishSponge, Spoon, Plate, Knife, DiningTable, Bowl, LaundryHamper, Vase, Stool, CoffeeTable, Poster, Bathtub, TissueBox, Footstool, BathtubBasin, ShowerCurtain, TVStand, Boots, RoomDecor, PaperTowelRoll, Ladle, Kettle, Safe, GarbageBag, TeddyBear, TableTopDecor, Dumbbell, Desktop, AluminumFoil, Window, LightSwitch, AppleSliced, BreadSliced, LettuceSliced, PotatoSliced, TomatoSliced\\
6. You can only pick up one object at a time. If the agent is holding an object, the agent should place or put down the object before attempting to pick up a second object.\\
7. Each object instance should instantiate a different InteractionObject class even if two object instances are the same object category. \\
Follow the output format provided earlier. Think step by step to carry out the instruction.

Write a Python script that could be executed by a household robot for the following:

dialogue: \{command\}

Python script: 
\end{tcolorbox}


\begin{tcolorbox}[
colback=white!10!white,
colframe=black!75!black,
title=AgentBench executor
prompt,
breakable]
\label{response-aug prompt agentbench}
Interact with a household to solve a task. Imagine you are an intelligent agent in a household environment and your target is to perform actions to complete the task goal. At the beginning of your interactions, you will be given the detailed description of the current environment and your goal to accomplish. For each of your turn, you will be given a list of actions which you can choose one to perform in this turn. You should choose from two actions: \"THOUGHT\" or \"ACTION\". If you choose \"THOUGHT\", you should first think about the current condition and plan for your future actions, and then output your action in this turn. Your output must strictly follow this format:\"THOUGHT: your thoughts.

ACTION: your next action

\"; If you choose \"ACTION\", you should directly output the action in this turn. Your output must strictly follow this format:\"ACTION: your next action
\". After your each turn, the environment will give you immediate feedback based on which you plan your next few steps. if the environment output \"Nothing happened\", that means the previous action is invalid and you should try more options.
Here is some experience you summarized before:
\{experience\} 

Reminder: 

1. the action must be chosen from the given available actions. Any actions except provided available actions will be regarded as illegal. 

2. Think when necessary, try to act directly more in the process.

"

\end{tcolorbox}

\section{Benchmark infromation}
\label{dataset}
\textbf{TEACh} The TEACh dataset \cite{padmakumar2022teach} is constructed on over 120 different AI2-THOR simulation environments \cite{kolve2017ai2} and encompasses more than 2000 embodied intelligence tasks aimed at completing household chores. These environments can be categorized into four hyper-environments: kitchen, living room, bedroom, and bathroom. The training set consists of 1482 data points, encompassing all four types of environments. The valid seen set is built with 181 data points across the four environments, with all simulation environments having appeared in the training set. In contrast, the valid unseen set is constructed with 612 data points in three types of environments: kitchen, living room, and bedroom, based on simulation environments that have not been previously encountered in the training set. Therefore, we consider the valid unseen set as out-of-domain (OOD) data and the valid seen set as in-domain (IND) data. Our tests are conducted on the Trajectory from Dialogue (TfD) benchmark \cite{padmakumar2022teach}, where the agent receives multiple rounds of interactive dialogue between a commander and a driver. The model must analyze the entire dialogue and make a series of decisions to complete all tasks mentioned in the dialogue.

\textbf{Alfworld} The Alfworld dataset \cite{shridhar2020alfworld} encompasses more than 4000 embodied intelligence tasks aimed at completing household chores. These tasks can be categorized into six hyper-task: "pick and place", "pick clean then place", "pick heat then place", "pick cool then place", "look at obj", and "pick two obj". We just select 20 successful experiences in each hyper-task. We use the AgentBench \cite{liu2023agentbench} for evaluation, it contains 20 data points in the dev set and 50 data points in the std set. Aligned with Alfworld, we consider the std set as out-of-domain (OOD) data and the dev set as in-domain (IND) data. 






\section{Prompt of Insight Generation}
\label{generator}
Below presents Pair-Mode Experience Generation Prompt and Success-Mode Experience Generation
Prompt. The parts with red are different. (For Alfworld, we just remove the part with "environment rules.")
\begin{tcolorbox}[
colback=white!10!white,
colframe=black!75!black,
title=Pair-Mode Insight Generation Prompt,
breakable]
You are an advanced reasoning agent that can add, edit, move or remove rules from your existing ruleset, based on forming new critiques of past task trajectories. The ruleset has three parts, GENERAL RULES, ENVIRONMENT RULES and TASK RULES. GENERAL RULES refers to rules that could used in all environment (Kitchens, LivingRooms, Bedrooms, and Bathrooms) and task. ENVIRONMENT RULES refers to rules that could used in all task in \{env\}. TASK RULES refers to rules that could used in a specific task. \textcolor{red}{You will be given two previous task trials with instruction:}
\\
\textcolor{red}{\{instruction\}\\
One trial is successful, and the other is unsuccessful. Here are the two previous trials to compare and critique:
\\
\\
Failed Trajectories:
\\
\{Failed Trajectories\}\\\\
Succeeded Trajectories:
\\
\{Succeeded Trajectories\}}
\\
Here are the EXISTING RULES:\\
GENERAL RULES:\\
\{general rules\}\\
ENVIRONMENT RULES:\\\{environment rules\}\\TASK RULES:\\\{task rules\}

\textcolor{red}{By examining and contrasting to the successful trial, and the list of existing rules, you can perform the following operations: add, edit, remove, move or agree so that the new rules are HIGH LEVEL critiques of the failed trial or proposed way of Thought in 3 parts, so they can be used to avoid similar failures when encountered with different questions in the future. Have an emphasis on critiquing how to perform better Thought and Action.}

Follow the below format:\\GENERAL RULES:\\<OPERATION> <RULE NUMBER> :<RULE>\\ENVIRONMENT RULES:\\<OPERATION> <RULE NUMBER> :<RULE>\\TASK RULES:\\<OPERATION> <RULE NUMBER> :<RULE>\\The rule number should increase between parts, for example if there is 4 general rules the first environment rule number should be 5.\\The available operations are: AGREE (if the existing rule is strongly relevant for the task), REMOVE(if one existing rule is contradictory or similar/duplicated to other existing rules), EDIT (if any existing rule is not general enough or can be enhanced, rewrite and improve it), ADD (add new rules that are very different from existing rules and relevant for other tasks.), MOVE(move rules between different level and reshape the rules if the rules are not general in all enviroment(for GENERAL RULES) or task(for GENERAL RULES or EMVIRONMENT RULES)). Each needs to CLOSELY follow their corresponding formatting below:\\AGREE <EXISTING RULE NUMBER>: <EXISTING RULE>

REMOVE <EXISTING RULE NUMBER>: <EXISTING RULE>

EDIT <EXISTING RULE NUMBER> :<NEW MODIFIED RULE>

ADD <NEW RULE NUMBER>: <NEW RULE>

MOVE <EXISTING RULE NUMBER>: <RESHAPED RULE>.(for example if you want to move a rule in environment rules with id 12 to task rules, you should use MOVE 12:<RESHAPED RULE> in task rules part)

Note1: MOVE command will remove the rules by number and add new rules in the part it present in and ADD command will add new rules in the part it present in.

Note2:If you believe some rules in general rule part can not be used in the \{env\}, you should just remove that rules instead of move it.

Note3:In task rules part, there may some task irrelevant with the trail now, DO NOT remove them

In the TASK RULES part, you should specify the task name in the <RULE> with the following format:<RULE CONTENT> (TASK: <TASK NAME>), the length of task name should be less than 20 characters and the number of task should less than 20. 

Do not mention the trials in the general rules because they should be GENERALLY APPLICABLE. Each rule should be concise and easy to follow.

Remember this robot can only generate python script. The execute subgoal and error log are gained from another robot which this robot can not communite. So each rules should focus on helping robot to plan and generate better python script to solve the question based on ONLY dialogue. And operation can be used MULTIPLE times. Do at most 4 operations in each parts (which means the max operation number in 3 parts is 4x3=12) and each existing rule can only get a maximum of 1 operation so just find the most important rules to operate. Do not operate rules in other parts.  Below are the operations you do to the above list of EXISTING RULES
\end{tcolorbox}

\begin{table*}[ht]
\centering
    \resizebox{1.0\textwidth}{!}{
    \begin{tabular}{lccccccccccc}
    \Xhline{1.5pt}
Insight source & 0 & 1 & 2 & 3 & 4 & 5 & 6 & 7 & 8 & 9 & 10 \\ \Xhline{1.5pt}
Expel & 14.29 & 1.19 & 16.67 & 23.81 & 13.1 & 2.38 & 7.14 & 14.29 & 7.14 & 0 & 0 \\ 
MSI Task & 0 & 0 & 6.42 & 8.26 & 12.84 & 1.83 & 11.93 & 30.28 & 19.27 & 4.59 & 4.59 \\
MSI General & 30.23 & 6.2 & 17.05 & 19.38 & 10.08 & 0 & 6.2 & 7.75 & 2.33 & 0.78 & 0 \\ \Xhline{1.5pt}
\end{tabular}
}
\caption{The insight's task-specific level under 3 sources. (0 for general insight and 10 for task-specific insight)}
\label{tab:data_insight}
\end{table*}

\begin{tcolorbox}[
colback=white!10!white,
colframe=black!75!black,
title=Success-Mode Insight Generation Prompt,
breakable]
You are an advanced reasoning agent that can add, edit, move or remove rules from your existing ruleset, based on forming new critiques of past task trajectories. The ruleset has three parts, GENERAL RULES, ENVIRONMENT RULES and TASK RULES. GENERAL RULES refers to rules that could used in all environment (Kitchens, LivingRooms, Bedrooms, and Bathrooms) and task. ENVIRONMENT RULES refers to rules that could used in all task in \{env\}. TASK RULES refers to rules that could used in a specific task. \textcolor{red}{You will be given successful task trials with instruction:\\\{instruction\}\\ Here are the trials:\\\{Succeeded Trajectories\}\\}
\\
Here are the EXISTING RULES:\\
GENERAL RULES:\\
\{general rules\}\\
ENVIRONMENT RULES:\\\{environment rules\}\\TASK RULES:\\\{task rules\}

\textcolor{red}{By examining the successful trials, and the list of existing rules, you can perform the following operations: add, edit, remove, move or agree so that the new rules are HIGH LEVEL insights of the successful trials or proposed way of Thought in 3 parts, so they can be used as helpful tips to different questions in the future. Have an emphasis on tips that help the agent perform better Thought and Action.\\}

Follow the below format:\\GENERAL RULES:\\<OPERATION> <RULE NUMBER> :<RULE>\\ENVIRONMENT RULES :\\<OPERATION> <RULE NUMBER> :<RULE>\\TASK RULES:\\<OPERATION> <RULE NUMBER> :<RULE>\\The rule number should increase between parts, for example if there is 4 general rules the first environment rule number should be 5.\\The available operations are: AGREE (if the existing rule is strongly relevant for the task), REMOVE(if one existing rule is contradictory or similar/duplicated to other existing rules), EDIT (if any existing rule is not general enough or can be enhanced, rewrite and improve it), ADD (add new rules that are very different from existing rules and relevant for other tasks.), MOVE(move rules between different level and reshape the rules if the rules are not general in all enviroment(for GENERAL RULES) or task(for GENERAL RULES or EMVIRONMENT RULES)). Each needs to CLOSELY follow their corresponding formatting below:\\AGREE <EXISTING RULE NUMBER>: <EXISTING RULE>

REMOVE <EXISTING RULE NUMBER>: <EXISTING RULE>

EDIT <EXISTING RULE NUMBER> :<NEW MODIFIED RULE>

ADD <NEW RULE NUMBER>: <NEW RULE>

MOVE <EXISTING RULE NUMBER>: <RESHAPED RULE>.(for example if you want to move a rule in environment rules with id 12 to task rules, you should use MOVE 12:<RESHAPED RULE> in task rules part)

Note1: MOVE command will remove the rules by number and add new rules in the part it present in and ADD command will add new rules in the part it present in.

Note2:If you believe some rules in general rule part can not be used in the \{env\}, you should just remove that rules instead of move it.

Note3:In task rules part, there may some task irrelevant with the trail now, DO NOT remove them

In the TASK RULES part, you should specify the task name in the <RULE> with the following format:<RULE CONTENT> (TASK: <TASK NAME>), the length of task name should be less than 20 characters and the number of task should less than 20. 

Do not mention the trials in the general rules because they should be GENERALLY APPLICABLE. Each rule should be concise and easy to follow.

Remember this robot can only generate python script. The execute subgoal and error log are gained from another robot which this robot can not communite. So each rules should focus on helping robot to plan and generate better python script to solve the question based on ONLY dialogue. And operation can be used MULTIPLE times. Do at most 4 operations in each parts (which means the max operation number in 3 parts is 4x3=12) and each existing rule can only get a maximum of 1 operation so just find the most important rules to operate. Do not operate rules in other parts.  Below are the operations you do to the above list of EXISTING RULES
\end{tcolorbox}

\section{Insight Selection Prompt}
\label{hash}
\begin{tcolorbox}[
colback=white!10!white,
colframe=black!75!black,
title=Insight Selection Prompt in Hashmap Index,
breakable]
You are a task selector trying to select task categories. \\
A household robot have just summarized some experience, and each experience belongs to a task category.\\
Now this robot is facing a new task, based on a dialogue between a <Driver> and a <Commander>, but this robot do not know which experience should be used in this task.\\
You should select task categories related to the task this robot facing.
You will be given a target task category, the target category is likely to be found in:\{task name\} \\
\\
Important: Your output should ONLY a list (categories seperated by commas) of the task categories from the list above.
\\
What are the task categories that related to \{dialogue\}?
\\
answer:

\end{tcolorbox}

\section{Example of Insight Selector}
\begin{tcolorbox}[
colback=white!10!white,
colframe=black!75!black,
title=Insight Selection Example]

\textbf{task}: put two soapbar in garbagecan \\ 
\textbf{selected subtask}: Object Placement, Distinguishing Similarities, Sequential Placement, Revealing Hidden Objects, Comprehensive Search

\end{tcolorbox}

\section{Insight High-Level Rate}
In the table \ref{tab:data_insight}, we compared the task-specific degree of three different insight sources in Alfworld, where 0 points are completely general (applicable to all tasks), 10 points are completely task-specific (can only be used for one specific task), and intermediate scores represent the degree to which they can be used for some tasks.

We have manually created three examples, each in the format:

(insight, thought, score).

For each example, the scores are respectively 0, 5, and 10. We have then asked the model (gpt-4-turbo-2024-04-09) to derive the score in a COT manner.

We can observe that the distribution of Expel is relatively uniform, the distribution of MSI Task tends to be around 7 points, while the distribution of MSI General leans towards 0-1 points.

This demonstrates that MSI indeed distinguishes between general insight and task-specific insight, and that task-specific insight is more targeted towards specific tasks.

\begin{tcolorbox}[
colback=white!10!white,
colframe=black!75!black,
title=Prompt of Rating Insight's Level]

prompt: You will be given an experience about houseworking, your task is to judge whether the experience is a general rule (all tasks in housework can be used) or a task-related rule. You should think step by step and give a score of 0-10, 0 means this experience is a general rule, and 10 means this experience is a task-related rule. Here are examples:
\end{tcolorbox}


\end{document}